\documentclass[runningheads]{llncs}
\pdfoutput=1

\usepackage{graphicx}

\usepackage{amsmath,amssymb,amsfonts}
\usepackage{textcomp}
\usepackage{xcolor}
\usepackage{xspace}
\usepackage{tabularx}
\usepackage{multirow}
\usepackage{algorithm}
\usepackage{algpseudocode}

\usepackage{footnote}
\makesavenoteenv{tabular}
\makesavenoteenv{table}
\usepackage{threeparttable}

\newcommand{\adv}{$\mathcal{A}$\xspace}
\newcommand{\victim}{$\mathcal{V}$\xspace}

\newcommand{\client}{API client\xspace}

\usepackage[hyphens,spaces,obeyspaces]{url}

\begin{document}
%
\title{Extraction of Complex DNN Models: Real Threat or Boogeyman?}
%
%
\author{Buse Gul Atli\inst{1} \and
Sebastian Szyller\inst{1}  \and
Mika Juuti\inst{2} \and
Samuel Marchal\inst{1,3} \and
N. Asokan\inst{2}}
%
\authorrunning{Buse G. Atli et al.}
%
\institute{Aalto University, 02150, Espoo Finland
\email{\{buse.atli,sebastian.szyller,samuel.marchal\}@aalto.fi} \\ \and
University of Waterloo, Waterloo, ON N2L 3G1, Canada  \\
\email{mika.juuti@uwaterloo.ca, asokan@acm.org} \\ \and
F-Secure Corporation, 00180 Helsinki, Finland
}
\maketitle              
\begin{abstract}
Recently, machine learning (ML) has introduced advanced solutions to many domains. Since ML models provide business advantage to model owners, protecting intellectual property of ML models has emerged as an important consideration.
Confidentiality of ML models can be protected by exposing them to clients only via prediction APIs.
However, model extraction attacks can steal the functionality of ML models using the information leaked to clients through the results returned via the API.
In this work, we question whether model extraction is a serious threat to complex, real-life ML models.
We evaluate the current state-of-the-art model extraction attack (Knockoff nets) against complex models.
We reproduce and confirm the results in the original paper. But we also show that the performance of this attack can be limited by several factors, including ML model architecture and the granularity of API response.
Furthermore, we introduce a defense based on distinguishing queries used for Knockoff nets from benign queries.
Despite the limitations of the Knockoff nets, we show that a more realistic adversary can effectively steal complex ML models and evade known defenses.

\end{abstract}
%
%
%
\section{Introduction}
\label{sec:introduction}

In recent years, machine learning (ML) has been applied to many areas with impressive results.
Use of ML models is now ubiquitous.
Major enterprises (Google, Apple, Facebook) utilize them in their products~\cite{techworld:2018}.
Companies gain business advantage by collecting proprietary data and training high quality models.
Hence, protecting the intellectual property embodied in ML models is necessary to preserve the business advantage of model owners.

Increased adoption of ML models and popularity of centrally hosted services led to the emergence of \textit{Prediction-As-a-Service} platforms.
Rather than distributing ML models to users, it is easier to run them on centralized servers having powerful computational resources and to expose them via \emph{prediction APIs}.
Prediction APIs are used to protect the confidentiality of ML models and allow for widespread availability of ML-based services that require users only to have an internet connection.
Even though users only have access to a prediction API, each response necessarily leaks some information about the model.
A \textit{model extraction attack}~\cite{tramer2016stealing} is one where an adversary (a malicious client) extracts information from a \emph{victim model} by frequently querying the model's prediction API. 
Queries and API responses are used to build a \emph{surrogate model} with comparable functionality and effectiveness. 
Deploying surrogate models deprive the model owner of its business advantage.
Many extraction attacks are effective against simple ML models~\cite{papernot2017practical,juuti2019prada} and defenses have been proposed against these simple attacks~\cite{lee2018defending,quiring2018forgottensib}.
However, extraction of complex ML models has got little attention to date.
Whether model extraction is a serious and realistic threat to real-life systems remains an open question.

Recently, a novel model extraction attack \emph{Knockoff nets}~\cite{orekondy2018knockoff} has been proposed against complex deep neural networks (DNNs). The paper reported empirical evaluations showing that Knockoff nets is effective at stealing \emph{any} image classification model.
This attack assumes that the adversary has access to
(a) \emph{pre}-\emph{trained} image classification models that are used as the basis for constructing the surrogate model,
(b) unlimited natural samples that are not drawn from the same distribution as the training data of the victim model
and (c) the full probability vector as the output of the prediction API.
Knockoff nets does not require the adversary to have any knowledge about the victim model, training data or its classification task (class semantics).
Although other and more recent model extraction attacks have been proposed~\cite{correia2018copycat,li2018query,jagielski2019high}, Knockoff nets remains the most effective one against complex DNN models with the weakest adversary model.
Moreover, there is no detection mechanism specifically tailored against model extraction attacks leveraging unlabeled natural data.
Therefore, the natural question is whether Knockoff nets is a realistic threat through extensive evaluation.

\vspace{-11pt}
\subsubsection{Goals and contributions.} Our goals are twofold in this paper.
First, we want to understand the conditions under which Knockoff nets constitutes a realistic threat.
Hence, we empirically evaluate the attack under different adversary models.
Second, we want to explore whether and under which conditions Knockoff nets can be mitigated or detected.
We claim the following contributions:
\begin{itemize}
	\item reproduce the empirical evaluation of Knockoff nets under its original adversary model to confirm that it can extract surrogate models exhibiting reasonable accuracy (53.5-94.8\%) for all five complex victim DNN models we built (Sect.~\ref{ssc:ko-eval}).
	\item introduce a defense, within the same adversary model, to detect Knockoff nets by differentiating in- and out-of-distribution queries (attacker's queries). This defense correctly detects up to 99\% of adversarial queries (Sect.~\ref{sec:defense}).
	\item revisit the original adversary model to investigate how the attack effectiveness changes with more realistic adversaries and victims (Section~\ref{ssec:victim-arch}).
	The attack effectiveness deteriorates when
	\begin{itemize}
		\item the adversary uses a model architecture for the surrogate that is different from that of the victim.
		\item the granularity of the victim's prediction API output is reduced (returning predicted class instead of a probability vector).
		\item the diversity of adversary queries is reduced.
	\end{itemize}
	On the other hand, the attack effectiveness can increase when the adversary has access to natural samples drawn from the \emph{same} distribution as the victim's training data. In this case, all existing attack detection techniques, including our own, are no longer applicable (Section~\ref{sec:st-improvements}).
\end{itemize}

\section{Background}
\label{sec:background}

\subsection{Deep Neural Networks}
\label{ssec:dnn}

A DNN is a function $F: \mathbb{R}^n \rightarrow \mathbb{R}^m$, where $n$ is the number of input features and $m$ is the number of output classes in a classification task. $F(x)$ gives a vector of length $m$ containing probabilities $p_j$ that input $x$ belongs to each class $c_j \in C$ for $j \in \lbrace1,m\rbrace$.
The predicted class, denoted $\hat{F}(x)$, is obtained by applying the \textit{argmax} function: $\hat{F}(x) = argmax(F(x))$.
$\hat{F}(x)$ tries to approximate a perfect oracle function $O_f: \mathbb{R}^n \rightarrow C$ which gives the true class $c$ for any input $x \in \mathbb{R}^n$.
The test accuracy $Acc(F)$ expresses the degree to which $F$ approximates $O_f$.

\subsection{Model Extraction Attacks}
\label{ssec:bg_extraction}

In a model extraction attack, the goal of an adversary \adv is to build a surrogate model $F_\mathcal{A}$ that imitates the model $F_\mathcal{V}$ of a victim \victim.
\adv wants to find an $F_\mathcal{A}$ having $Acc(F_\mathcal{A})$ as close as possible to $Acc(F_\mathcal{V})$ on a test set.
\adv builds its own dataset $D_{\mathcal{A}}$ and implements the attack by sending queries to the prediction API of $F_\mathcal{V}$ and obtaining predictions $F_\mathcal{V}(x)$ for each query $x$, where $\forall x \in D_{\mathcal{A}}$.
\adv uses the \emph{transfer set} $\{D_{\mathcal{A}}, F_\mathcal{V}(D_{\mathcal{A}})\}$ to train a surrogate model $F_\mathcal{A}$.

According to prior work on model extraction~\cite{papernot2017practical,juuti2019prada}, we can divide \adv's capabilities into three categories: \textit{victim model knowledge}, \textit{data access}, \textit{querying strategy}.

\vspace{-11pt}
\subsubsection{Victim model knowledge.} Model extraction attacks operate in a \emph{black-box} setting. \adv does not have access to model parameters of $F_\mathcal{V}$ but can query the prediction API without any limitation on the number of queries. \adv might know the exact architecture of $F_\mathcal{V}$, its hyperparameters or its training process.
Given the purpose of the API (e.g., image recognition) and expected complexity of the task, \adv may attempt to guess the architecture of $F_\mathcal{V}$~\cite{papernot2017practical}.  $F_\mathcal{V}$'s prediction API may return one of the following: the probability vector, top-k labels with confidence scores or only the predicted class.

\vspace{-11pt}
\subsubsection{Data access.} Possible capabilities of \adv for data access vary in different model extraction attacks. \adv can have access to a small subset of natural samples from \victim's training dataset~\cite{papernot2017practical,juuti2019prada}. \adv may not have access to \victim's training dataset $D_{\mathcal{V}}$ but may know the ``domain'' of data and have access to natural samples that are close to \victim's training data distribution (e.g., images of dogs in the task of identifying dog breeds)~\cite{correia2018copycat}. \adv can use widely available natural samples that are different from \victim's training data distribution~\cite{orekondy2018knockoff}. Finally, \adv can construct $D_{\mathcal{A}}$ with only synthetically crafted samples~\cite{tramer2016stealing}.

\vspace{-11pt}
\subsubsection{Querying strategy.} Querying is the process of submitting a sample to the prediction API. If \adv relies on synthetic data, it crafts samples that would help it train $F_\mathcal{A}$ iteratively. Otherwise, \adv first collects its samples $D_{\mathcal{A}}$, queries the prediction API with the complete $D_{\mathcal{A}}$, and then trains the surrogate model with $\{D_{\mathcal{A}}, F_\mathcal{V}(D_{\mathcal{A}})\}$.

\section{Knockoff Nets Model Extraction Attack}
\label{sec:knockoff}
In this section, we study the Knockoff nets model extraction attack~\cite{orekondy2018knockoff} which achieves state-of-the-art
performance against complex DNN models. Knockoff nets works without access to \victim's training data distribution, model architecture and classification task. 

\subsection{Attack Description}
\label{ssec:knockoff_description}
\subsubsection{Adversary model.}\label{ssec:adversary_model} The goal of \adv is \textit{model functionality stealing}~\cite{orekondy2018knockoff}: \adv wants to train a surrogate model $F_\mathcal{A}$ that performs similarly on a classification task for which prediction API's $F_\mathcal{V}$ was designed.
\adv has no information about $F_\mathcal{V}$ including model architecture, internal parameters and hyperparameters. Moreover, \adv does not have access to \victim's training data, prediction API's purpose or output class semantics. \adv is a weaker adversary than previous work in~\cite{papernot2017practical,juuti2019prada} due to these assumptions.
However, \adv can collect an unlimited amount of varied real-world data from online databases 
and can query prediction API without any constraint on the number of queries.
API always returns a complete probability vector as an output for each legitimate query.
\adv is not constrained in memory and computational capabilities and uses publicly available pre-trained complex DNN models as a basis for $F_\mathcal{A}$~\cite{kornblith2018better}.

\vspace{-11pt}
\subsubsection{Attack strategy.} \adv first collects natural data from online databases for constructing unlabeled dataset $D_\mathcal{A}$.
For each query $x$, $\forall x \in D_\mathcal{A}$, \adv obtains a complete probability vector $F_\mathcal{V}(x)$ from the prediction API.
\adv uses this transfer set $\{D_{\mathcal{A}}, F_\mathcal{V}(D_{\mathcal{A}})\}$
to repurpose learned features of a complex pre-trained model with transfer learning~\cite{kornblith2019better}.
In the Knockoff nets setting,
\victim offers image classification and \adv constructs $D_{\mathcal{A}}$ by sampling a subset of the ImageNet dataset~\cite{deng2009imagenet}.

\subsection{Knockoff Nets: Evaluation}
\label{ssc:ko-eval}

We first implement Knockoff nets under the original adversary model explained in Section~\ref{ssec:knockoff_description}.
We use the datasets and experimental setup described in~\cite{orekondy2018knockoff} for constructing both $F_{\mathcal{V}}$  and $F_{\mathcal{A}}$ .
We also evaluate two additional datasets to contrast our results with previous work.

\subsubsection{Datasets.}\label{sssc:datasets} We use Caltech~\cite{griffin2007caltech}, CUBS~\cite{WelinderEtal2010} and Diabetic Retinopathy (Diabetic5)~\cite{kaggle:Diabetic} datasets as in~\cite{orekondy2018knockoff} for training $F_\mathcal{V}$'s and reproduce experiments where Knockoff nets was successful. Caltech is composed of various images belonging to 256 different categories.
CUBS contains images of 200 bird species and is used for fine-grained image classification tasks. Diabetic5 contains high-resolution retina images labeled with five different classes indicating the presence of diabetic retinopathy.
We augment Diabetic5 using preprocessing techniques recommended in \footnote{\url{https://github.com/gregwchase/dsi-capstone}} to address the class imbalance problem.
For constructing $D_{\mathcal{A}}$, we use a subset of ImageNet, which contains 1.2M images belonging to 1000 different categories.  $D_{\mathcal{A}}$ includes randomly sampled 100,000 images from Imagenet, 100 images per class. 42\% of labels in Caltech and 1\% in CUBS are also present in ImageNet.
There is no overlap between Diabetic5 and ImageNet labels.

Additionally, we use CIFAR10~\cite{krizhevsky2009cifar10}, depicting animals and vehicles divided into 10 classes, and GTSRB~\cite{stallkamp2011german}, a traffic sign dataset with 43 classes. CIFAR10 contains broad, high level classes while GTSRB contains domain specific and detailed classes. These datasets do not overlap with ImageNet labels and they were partly used in prior model extraction work~\cite{papernot2017practical,juuti2019prada}. 
We resize images with bilinear interpolation, where applicable.

All datasets are divided into training and test sets and summarized in Table~\ref{tab:datasets}. All images in both training and test sets are normalized with mean and standard deviation statistics specific to ImageNet.

\begin{table}[htb]
\centering
	\caption{Image datasets used to evaluate Knockoff nets. GTSRB and CIFAR10 are resized with bilinear interpolation before training pre-trained classifiers.}
	\label{tab:datasets}
    \centering
\begin{tabular}{ p{\dimexpr 0.20\linewidth-2\tabcolsep}
						  p{\dimexpr 0.20\linewidth-2\tabcolsep}
                        p{\dimexpr 0.20\linewidth-2\tabcolsep}
                        p{\dimexpr 0.20\linewidth-2\tabcolsep}
                        p{\dimexpr 0.20\linewidth-2\tabcolsep}}
   \hline \hfil \multirow{2}{*}{Dataset} & \hfil \multirow{2}{*}{Image size}& \hfil \multirow{2}{*}{Num. of Classes}	& \multicolumn{2}{c}{Number of samples} \\ 
			&  & 	& \hfil Train		& \hfil Test \\ \hline
		\hfil Caltech 	& \hfil  224x224				& \hfil 256 & \hfil 23,703 & \hfil 6,904 \\
		\hfil CUBS       & \hfil  224X224            & \hfil 200 & \hfil 5994    & \hfil 5794 \\
		\hfil Diabetic5 & \hfil 224x224             & \hfil 5     & \hfil 85,108 & \hfil 21,278 \\
		\hfil GTSRB		& \hfil 32x32 / 224x224	& \hfil 43	  & \hfil 39,209 & \hfil 12,630 \\
		\hfil CIFAR10	& \hfil 32x32 / 224x224	& \hfil 10	  & \hfil 50,000 & \hfil 10,000 \\ \hline
	\end{tabular}
\end{table}

\vspace{-11pt}
\subsubsection{Training victim models.}\label{sssc:victimmodels} To obtain complex victim models, we fine-tune weights of a pre-trained ResNet34~\cite{he2016deep} model. We train 5 complex victim models using the datasets summarized in Table~\ref{tab:datasets} and name these victim models \{Dataset name\}-RN34. In training, we use SGD optimizer with an initial learning rate of 0.1 that is decreased by a factor of 10 every 60 epochs over 200 epochs.

\vspace{-11pt}
\subsubsection{Training surrogate models.}\label{sssc:surrogatemodels} To build surrogate models, we fine-tune weights of a pre-trained ResNet34~\cite{he2016deep} model.
We query $F_{\mathcal{V}}$'s prediction API with samples from $D_{\mathcal{A}}$ and obtain $\{D_{\mathcal{A}}, F_\mathcal{V}(D_{\mathcal{A}})\}$. We train surrogate models using an SGD optimizer with an initial learning rate of 0.01 that is decreased by a factor of 10 every 60 epochs over 100 epochs. We use the same model architecture for both $F_{\mathcal{V}}$ and $F_{\mathcal{A}}$ in order to replicate the experiments in the original paper. Additionally, we discuss the effect of model architecture mismatch in Section~\ref{ssec:victim-arch}.

\vspace{-11pt}
\subsubsection{Experimental results.}\label{ssc:results} Table~\ref{tab:knockoff_results} presents the test accuracy of $F_\mathcal{V}$ and $F_\mathcal{A}$ in our reproduction as well as experimental results reported in the original paper ($F_\mathcal{V_{\dagger}}, F_\mathcal{A_{\dagger}}$). The attack effectiveness
against Caltech-RN34 and CUBS-RN34 models is consistent with the corresponding values reported in~\cite{orekondy2018knockoff}. We found that $F_\mathcal{A}$ against Diabetic5-RN34 does not recover the same degree of performance. This inconsistency is a result of different transfer sets labeled by two different $F_\mathcal{V}$'s.

As shown in Table~\ref{tab:knockoff_results}, Knockoff nets is effective against pre-trained complex DNN models. Knockoff nets can imitate the functionality of $F_\mathcal{V}$ via \adv's transfer set, even though $D_{\mathcal{A}}$ is completely different from \victim's training data. We will discuss the effect of transfer set with more detail in Section~\ref{sssc:trainingdata}.

\begin{table}[htb]
	\centering
		\caption{Test accuracy $Acc(\cdot)$ of $F_\mathcal{V}$, $F_\mathcal{A}$ in our reproduction and $F_\mathcal{V_{\dagger}}$, $F_\mathcal{A_{\dagger}}$ reported by~\cite{orekondy2018knockoff}. Good surrogate models are in bold based on their performance recovery ($ Acc(F_\mathcal{A})/Acc(F_\mathcal{V}) \times $).}
		\label{tab:knockoff_results}
			\begin{tabular}{p{\dimexpr 0.22\linewidth-2\tabcolsep}
						  p{\dimexpr 0.16\linewidth-2\tabcolsep}
                        p{\dimexpr 0.22\linewidth-2\tabcolsep}
                        p{\dimexpr 0.16\linewidth-2\tabcolsep}
                        p{\dimexpr 0.22\linewidth-2\tabcolsep}} \hline
				\hfil $F_\mathcal{V}$	& \hfil $Acc(F_\mathcal{V})$   &  \hfil $Acc(F_\mathcal{A})$ & \hfil $Acc(F_\mathcal{V_{\dagger}})$ & \hfil $Acc(F_\mathcal{A_{\dagger}})$  \\ \hline
				\hfil  Caltech-RN34   & \hfil 74.6\%   & \hfil 72.2\% (\textbf{0.97} $\times$) &  \hfil 78.8\% & \hfil 75.4\% (\textbf{0.96 $\times$}) \\
				\hfil CUBS-RN34       & \hfil 77.2\%   & \hfil 70.9\% (0.91 $\times$)   & \hfil 76.5\%   &  \hfil 68.0\% (0.89 $\times$) \\
				\hfil  Diabetic5-RN34 & \hfil 71.1\%   & \hfil 53.5\% (0.75 $\times$)   & \hfil 58.1\%   &  \hfil 47.7\% (0.82 $\times$)  \\
				\hfil  GTSRB-RN34     & \hfil 98.1\%   & \hfil 94.8\% (\textbf{0.97} $\times$) &  \hfil - &\hfil   -\\
				\hfil CIFAR10-RN34   & \hfil 94.6\%   & \hfil 88.2\% (\textbf{0.93} $\times$) &  \hfil - & \hfil  -\\ \hline
			\end{tabular}
\end{table}

\section{Detection of Knockoff Nets Attack}
\label{sec:defense}

In this section, we present a method designed to detect queries used for Knockoff nets.
We analyze attack effectiveness w.r.t. the capacity of the model used for detection and the overlap between \adv's and \victim's training data distributions.
Finally, we investigate attack effectiveness when \adv's queries are detected and additional countermeasures are taken.

\subsection{Goals and Overview}\label{ssec:def-goals}
DNNs are trained using datasets that come from a specific distribution $\mathcal{D}$.
Many benchmark datasets display distinct characteristics that make them identifiable (e.g. cars in CIFAR10 vs ImageNet) as opposed to being representative of real-world data~\cite{torralba2011datasets}.
A DNN trained using such data might be overconfident, i.e. it gives wrong predictions with high confidence scores, when it is evaluated with samples drawn from a different distribution $\mathcal{D}'$. 
Predictive uncertainty is unavoidable when a DNN model is deployed for use via a prediction API. In this case, estimating predictive uncertainty is crucial to reduce over-confidence and provide better generalization for unseen samples.
Several methods were
introduced~\cite{hendrycks17baseline,liang2017principled,lee2018simple} to measure the predictive uncertainty by detecting out-of-distribution samples in the domain of image recognition.
Baseline~\cite{hendrycks17baseline} and ODIN~\cite{liang2017principled} methods analyze the softmax probability distribution of the DNN to identify out-of-distribution samples.
A recent state-of-the-art-method~\cite{lee2018simple} detects out-of-distribution samples based on their Mahalanobis distance~\cite{bishop2006pattern} to the closest class-conditional distribution.
Although these methods were tested against adversarial samples in evasion attacks, their detection performance against Knockoff nets is unknown.
What is more, their performance heavily relies on the choice of threshold value which corresponds to the rate of correctly identified in-distribution samples (TNR rate).

Our goal is to detect queries that do not correspond to the main classification task of \victim's model.
In case of Knockoff nets, this translates to identifying inputs that come from a different distribution than \victim's training set.
Queries containing such images constitute the distinctive aspect of the adversary model in Knockoff nets: 1) availability of large amount of unlabeled data 2) limited information about the purpose of the API. To achieve this, we propose a binary classifier 
based on the ResNet architecture.
It differentiates inputs from and out of \victim's data distribution.
Our binary classifier requires a labeled out-of-distribution dataset for the training phase unlike other
detection methods~\cite{hendrycks17baseline,liang2017principled,lee2018simple}.
However, binary classifiers have higher accuracy in the absence of adversary's queries \cite{biggio2015oneandhalf} than one-class classifiers used in the prior work.

Our solution can be used as a filter placed in front of the prediction API.

\subsection{Training Setup}\label{ssec:def-setup}

\subsubsection{Datasets.}\label{sssec:def-datasets} When evaluating our method, we consider all $F_\mathcal{V}$'s we built before in Section~\ref{ssc:ko-eval}. To train our binary classifiers, we combine \victim's training samples that are used to build $F_\mathcal{V}$ (in-distribution) and 90,000 randomly sampled images from ImageNet (out-distribution), 90 images per class. To construct a test dataset, we combine \victim's corresponding test samples and another subset of ImageNet containing 10,000 samples, 10 images per class. ImageNet serves the purpose of a varied and easily available dataset that \adv could use. We assume that legitimate clients query the prediction API with in-distribution test samples and \adv queries it with 10,000 ImageNet samples. In order to measure the generalizability of our detector, we also consider the case where \adv queries the prediction API with 20,000 samples from the
OpenImages~\cite{kuznetsova2018open} dataset that does not overlap with ImageNet. We use the same test datasets to measure the performance of other state-of-the-art out-of-distribution detectors.

\vspace{-11pt}
\subsubsection{Training binary classifer.}\label{sssec:def-training} In our experiments, we examine two types of models: 1) ResNet models trained from scratch and 2) pre-trained ResNet models with frozen weights where we replace the final layer with binary logistic regression.
In this section, we refer to different ResNet models as RN followed by the number indicating the number of layers, e.g. RN34; we further use the LR suffix to highlight pre-trained models with a logistic regression layer.

We assign label $0$ to in-distribution samples (\victim's dataset) and $1$ to out-of-distribution samples (90,000 ImageNet samples). All images are normalized according to ImageNet-derived mean and standard deviation.
We apply the same labeling and normalization procedure to the \adv's transfer sets
(both 10,000 ImageNet and 20,000 OpenImages samples).
To train models from scratch (models RN18 and RN34), we use the ADAM optimizer~\cite{kingma2014adam} with initial learning rate of 0.001 for the first 100 epochs and 0.0005 for the remaining 100 (200 total).
Additionally, we repeat the same training procedure while removing images whose class labels overlap with ImageNet from \victim's dataset (models RN18*, RN34*, RN18*LR, RN34*LR, RN101*LR, RN152*LR).
This will minimize the risk of false positives and simulate the scenario with no overlap between the datasets.
To train models with the logistic regression layer (models RN18*LR, RN34*LR, RN101*LR, RN152*LR), we take ResNet models pre-trained on ImageNet.
We replace the last layer with a logistic regression model and freeze the remaining layers.
We train logistic regression using the LBFGS solver~\cite{zhu94lbfgs} with L2 regularization and use 10-fold cross-validation to find the optimal value of the regularization parameter.

\subsection{Experimental Results}\label{ssec:def-results}
We divide our experiments into two phases.
In the first phase, we select CUBS and train binary classifiers with different architectures in order to identify the optimal classifier.
We assess the results using the rate of correctly detected in- (true negative rate, TNR) and out-of-distribution samples (true positive rate, TPR).
In the second phase, we evaluate the performance of the selected optimal architecture using all datasets in Section~\ref{sssc:datasets} and assess it based on the achieved TPR and TNR.

\begin{table}[htb]
	\centering
	\caption{Distinguishing \adv's ImageNet transfer set (TPR) from in-distribution samples corresponding to CUBS test set (TNR). Results are reported for models trained from scratch (RN18, RN34), trained from scratch excluding overlapping classes (models RN18*, RN34*) and using pre-trained models with logistic regression (models RN18*LR, RN34*LR, RN101*LR, RN152*LR. Best results are in bold.}
	\label{tab:detection-architecture}
	\begin{tabular}{p{\dimexpr 0.24\linewidth-2\tabcolsep}
						  p{\dimexpr 0.19\linewidth-2\tabcolsep}
                        p{\dimexpr 0.19\linewidth-2\tabcolsep}
                        p{\dimexpr 0.19\linewidth-2\tabcolsep}
                        p{\dimexpr 0.19\linewidth-2\tabcolsep}}\hline
		 & \multicolumn{4}{c}{Model} \\
		 & \hfil RN18  & \hfil RN34  & \hfil RN18* & \hfil RN34* \\ \hline
		 \hfil TPR/TNR & \hfil 86\% / 83\% & \hfil 94\% / 80\% & \hfil 90\% / 83\% & \hfil \textbf{95}\% / 82\% \\ \hline
		 & \multicolumn{4}{c}{Model} \\
		  & \hfil RN18*LR  & \hfil RN34*LR   & \hfil RN101*LR & \hfil RN152*LR  \\ \hline
		  \hfil TPR/TNR &  \hfil 84\% / 84\% & \hfil 93\% / 89\% & \hfil 93\% / \textbf{93}\% & \hfil 93\% / \textbf{93}\%
		 \\ \hline
	\end{tabular}
\end{table}

\begin{table}[htb]
	\caption{Distinguishing in-distribution test samples from \adv's transfer set as out-of-distribution samples. Comparison of our method with Baseline~\cite{hendrycks17baseline}, ODIN~\cite{liang2017principled} and Mahalanobis~\cite{lee2018simple} w.r.t TPR (correctly detected out-of-distribution samples) and TNR (correctly detected in-distribution samples). Best results are in bold.}
	\label{tab:detection-results}
			\centering
			\begin{tabular}{p{\dimexpr 0.17\linewidth-2\tabcolsep}
						  p{\dimexpr 0.15\linewidth-2\tabcolsep}
                        p{\dimexpr 0.09\linewidth-2\tabcolsep}
                        p{\dimexpr 0.09\linewidth-2\tabcolsep}
                        p{\dimexpr 0.26\linewidth-2\tabcolsep}
                         p{\dimexpr 0.26\linewidth-2\tabcolsep}} \hline
			\hfil \adv's	& \hfil In-dist.		& \multicolumn{2}{c}{Ours}  	& \multicolumn{2}{c}{Baseline / ODIN / Mahalanobis}  \\
			\hfil transfer set & \hfil dataset & TPR 	& TNR 	& TPR (at TNR Ours) & TPR (at TNR 95\%) \\ \hline
			\hfil \multirow{5}{*}{ImageNet}
			& \hfil Caltech 	    & \hfil 63\% & \hfil 56\% 	& \hfil 87\% / \textbf{88}\% / 59\%	& \hfil 13\% / 11\% / 5\%\text{ } \\
			& \hfil CUBS 		& \hfil \textbf{93}\% & \hfil \textbf{93}\% &
			\hfil 48\% / 54\% / 19\%  & \hfil 39\% / 43\% / 12\% \\
			& \hfil Diabetic5	& \hfil \textbf{99}\% & \hfil \textbf{99}\% & \hfil \text{ }1\%  / 25\% / 98\%   & \hfil \text{ }5\% / 49\% / \textbf{99}\% \\
			& \hfil GTSRB 		& \hfil \textbf{99}\% & \hfil \textbf{99}\% & \hfil 42\% / 56\% / 71\%   & \hfil 77\% / 94\% / 89\% \\
			& \hfil CIFAR10 	& \hfil \textbf{96}\% & \hfil \textbf{96}\% & \hfil 28\% / 54\% / 89\%	& \hfil 33\% / 60\% / 91\% \\ \hline
			\hfil \multirow{5}{*}{OpenImages}
			& \hfil Caltech 	   & \hfil 61\% & \hfil 59\% &
			\hfil \textbf{83}\% / \textbf{83}\% / 6\%\text{ } &
			\hfil 11\% / 11\% / 6\%\text{ } \\
			&	\hfil CUBS 	   & \hfil \textbf{93}\% & \hfil \textbf{93}\% &
			\hfil 47\% / 50\% / 14\%   & \hfil 37\% / 44\% / 14\% \\
			&	\hfil Diabetic5	& \hfil \textbf{99}\% & \hfil \textbf{99}\% & \hfil \text{ }1\% / 21\% / \textbf{99}\%  	& \hfil \text{ }4\% / 44\% / \textbf{99}\% \\
			&	\hfil	GTSRB 		& \hfil \textbf{99}\% & \hfil \textbf{99}\%
			& \hfil 44\% / 64\% / 75\%  & \hfil 76\% / 93\% / 87\% \\
			&	\hfil CIFAR10 	& \hfil \textbf{96}\% & \hfil \textbf{96}\% & \hfil 27\% / 56\% / 92\%	& \hfil 33\% / 62\% / 95\% \\ \hline
			\end{tabular}
\end{table}

As presented in the Table~\ref{tab:detection-architecture}, we find that the optimal architecture is RN101*LR: pre-trained ResNet101 model with logistic regression replacing the final layer.
Table~\ref{tab:detection-architecture} also shows that increasing model capacity improves detection accuracy. 
For the remaining experiments we use RN101*LR since it achieves the same TPR and TNR as RN152*LR while being faster in inference.

Prior work~\cite{yosinski2014transferable,kornblith2019better}
has shown that pre-trained DNN features transfer better when tasks are similar.
In our case, half of task is identical to the pre-trained task
(recognizing ImageNet images).
Thus it might be ideal to avoid modifying
network parameters and keep pre-trained model parameters frozen by replacing
the last layer with a logistic regression.
Another benefit of using logistic regression over complete fine-tuning
is that
pre-trained embeddings can be \emph{pre-calculated once} at a negligible cost,
after which training can proceed without performance penalties on CPU in
a matter or minutes.
Thus, model owners can cheaply train an effective model extraction defense.
Such a defense can have wide applicability for small-scale and medium-scale model owners.
Finally, since our defense mechanism is stateless, it does not depend on prior queries made by the adversary nor does it keep other state. It handles each query in isolation; therefore, it can not be circumvented by sybil attacks.

Maintaining high TNR is important for usability reasons.
Table~\ref{tab:detection-results} showcases results for all datasets.
We compare our approach with existing state-of-the-art methods
detecting out-of-distribution samples \footnote{\url{https://github.com/pokaxpoka/deep_Mahalanobis_detector}} when they are also deployed to identify \adv's queries. We report results for these methods with optimal hyper-parameters (c.f. Table~\ref{tab:detection-results}).
Note that other methods are threshold-based detectors, they require setting TNR to a value before detecting \adv's queries. 
Our method achieves high TPR ($>90\%$) on all \victim's but Caltech-RN34 and very high ($>99\%$) for GTSRB-RN34 and Diabetic5-RN34.
Furthermore, our method outperforms other state-of-the-art approaches when detecting \adv's queries. 
These results are consistent considering the overlap between \victim's training dataset and our subsets of
ImageNet and OpenImages (\adv's transfer set).
GTSRB and Diabetic5 have no overlap with ImageNet or OpenImages.
On the other hand, CUBS, CIFAR10 and Caltech contain images that represent either the same classes or families of classes (as in CIFAR10) as ImageNet and OpenImages.
This phenomena is particularly pronounced in case of Caltech which has strong similarities to ImageNet and OpenImages.
While TPR remains significantly above the random 50\%, such a model is not suitable for deployment.
Although other methods can achieve higher TPR on Caltech (87-88\%), we measured this value with TNR fixed at 56\%.
All models fail to discriminate Caltech samples from \adv's queries when constrained to have a more reasonable TNR 95\%. We find that
our defense method  works better with prediction APIs that have specific tasks (such as traffic sign recognition), as opposed to general purpose classifiers that can classify thousands of fine-grained classes. We will discuss how a more realistic \adv can evade these detection mechanisms in Section~\ref{sec:st-improvements}.
\section{Revisiting the Adversary Model}
\label{sec:revisiting}

We aim to identify capabilities and limitations of Knockoff nets under different experimental setups with more realistic assumptions.
We evaluate Knockoff nets when 1) $F_\mathcal{A}$ ad $F_\mathcal{V}$ have completely different architectures, 2) the granularity of $F_{\mathcal{V}}$'s prediction API output changes, and 3) \adv can access data closer to \victim's training data distribution.
We also discuss $D_{\mathcal{A}}$'s effect on the surrogate model performance.

\subsection{Victim Model Architecture}\label{ssec:victim-arch}

We measure the performance of Knockoff nets when $F_{\mathcal{V}}$ does not use pre-trained DNN model but is trained from scratch with a completely different architecture for its task.
We apply 5-layer GTSRB-5L and 9-layer CIFAR10-9L $F_\mathcal{V}$'s as described in previous model extraction work~\cite{juuti2019prada}.
These models are trained using Adam optimizer with learning rate of 0.001 that is decreased to 0.0005 after 100 epochs over 200 epochs.
The training procedure of surrogate models is the same as in Section~\ref{sssc:surrogatemodels}.
Thus, GTSRB-5L and CIFAR10-9L have different architectures and optimization algorithms than those used by \adv.
As shown in Table~\ref{tab:victim_architecture}, Knockoff nets performs well when both $ F_\mathcal{V}$ and $ F_\mathcal{A}$ use pre-trained models even if \adv uses a different pre-trained model architecture 
(VGG16~\cite{simonyan2014very}). However, the attack effectiveness decreases when $ F_\mathcal{V}$ is specifically designed for the given task and does not base its performance on any pre-trained model.

\begin{table}[htb]
	\centering
		\caption{Test accuracy $Acc(\cdot)$ of $F_\mathcal{V}$, $F_{\mathcal{A}_{R}}$ and $F_{\mathcal{A}_{V}}$
		and the performance recovery of surrogate models. $\mathcal{A}_{R}$ uses ResNet34 and $\mathcal{A}_{V}$ uses VGG16 for surrogate model architecture.}
		\label{tab:victim_architecture}
		\begin{tabular}{p{\dimexpr 0.20\linewidth-2\tabcolsep}
						    p{\dimexpr 0.20\linewidth-2\tabcolsep}
                          p{\dimexpr 0.25\linewidth-2\tabcolsep}
                          p{\dimexpr 0.25\linewidth-2\tabcolsep}} \hline
			\hfil $F_\mathcal{V}$	& \hfil $Acc(F_\mathcal{V})$   &  \hfil $Acc(F_{\mathcal{A}_{R}})$ & \hfil $Acc(F_{\mathcal{A}_{V}})$\\ \hline
			\hfil GTSRB-RN34     & \hfil 98.1\%   &
			\hfil 94.8\% (0.97 $\times$)  &  \hfil 90.1 (0.92 $\times$) \\
			\hfil GTSRB-5L       & \hfil 91.5\%   &
			\hfil 54.5\% (0.59 $\times$)  &\hfil 56.2 (0.61 $\times$) \\
			\hfil CIFAR10-RN34   & \hfil 94.6\%   &
			\hfil 88.2\% (0.93 $\times$)  &  \hfil 82.9 (0.87 $\times$)  \\
			\hfil CIFAR10-9L     & \hfil 84.5\%   & 
			\hfil 61.4\% (0.73 $\times $)  &  \hfil 64.7 (0.76 $\times$) \\ \hline
		\end{tabular}
\end{table}

\subsection{Granularity of Prediction API Output}\label{ssec:victim-out}
If $F_\mathcal{V}$'s prediction API gives only the predicted class or truncated results, such as top-k predictions or rounded version of the full probability vector for each query, performance of the surrogate model degrades.
Table~\ref{tab:different_voutput} shows this limitation, where the prediction API gives complete probability vector to $\mathcal{A}_{p}$ and only predicted class to $\mathcal{A}_{c}$. Table~\ref{tab:different_voutput} also demonstrates that the amount of degradation is related to the number of classes in $F_\mathcal{V}$, since \adv obtains comparatively less information if the actual number of classes is high and the granularity of response is low. For example, the degradation is severe when Knockoff nets is implemented against Caltech-RN34 and CUBS-RN34 having more than or equal to 200 classes. However, degradation is low or zero when Knockoff nets is implemented against other models (Diabetic5-RN34, GTSRB-RN34, CIFAR10-RN34).

\begin{table}[htb]
	\centering
	\caption{Test accuracy $Acc(\cdot)$ of $F_\mathcal{V}$, $F_{\mathcal{A}_{p}}$, $F_{\mathcal{A}_{c}}$ and the performance recovery of surrogate models. $\mathcal{A}_{p}$ receives complete probability vector and $\mathcal{A}_{c}$ only receives predicted class from the prediction API.}  
	\label{tab:different_voutput}
		\begin{tabular}{p{\dimexpr 0.36\linewidth-2\tabcolsep}
						    p{\dimexpr 0.20\linewidth-2\tabcolsep}
                          p{\dimexpr 0.22\linewidth-2\tabcolsep}
                          p{\dimexpr 0.22\linewidth-2\tabcolsep}} \hline
			\hfil $F_\mathcal{V}$	& \hfil $Acc(F_\mathcal{V})$   & \hfil $Acc(F_{\mathcal{A}_{p}})$  & \hfil $Acc(F_{\mathcal{A}_{c}})$  \\ \hline
			\hfil Caltech-RN34 (256 classes)  & \hfil 74.6\%  & \hfil 68.5\% (0.92 $\times$)  & \hfil 41.9\%
			(0.56 $\times$)  \\
			\hfil CUBS-RN34  (200 classes)    & \hfil 77.2\%  &  \hfil 54.8\% (0.71 $\times$)    & \hfil 18.0\%
			(0.23 $\times$)  \\
			\hfil Diabetic5-RN34 (5 classes) & \hfil 71.1\%  & \hfil 59.3\% (0.83 $\times$)   & \hfil 54.7\%
			(0.77 $\times$)  \\
			\hfil GTSRB-RN34 (43 classes)    & \hfil 98.1\%  &  \hfil 92.4\% (0.94 $\times$)   & \hfil 91.6\% (0.93 $\times$)    \\
			\hfil CIFAR10-RN34 (10 classes)   & \hfil 94.6\%  &  \hfil 71.1\% (0.75 $\times$)   & \hfil 53.6\% (0.57 $\times$)  \\			\hline
		\end{tabular}
	\end{table}

Many commercial prediction APIs return top-k outputs for queries (Clarifai returns top-10 outputs and Google Cloud Vision returns up to top-20 outputs from more than 10000 labels). Therefore, attack effectiveness will likely degrade when it is implemented against such real-world prediction APIs.

\subsection{Transfer Set Construction}\label{sssc:trainingdata}

\begin{figure}[htb]
	\centering
	\includegraphics[width=0.42\linewidth]{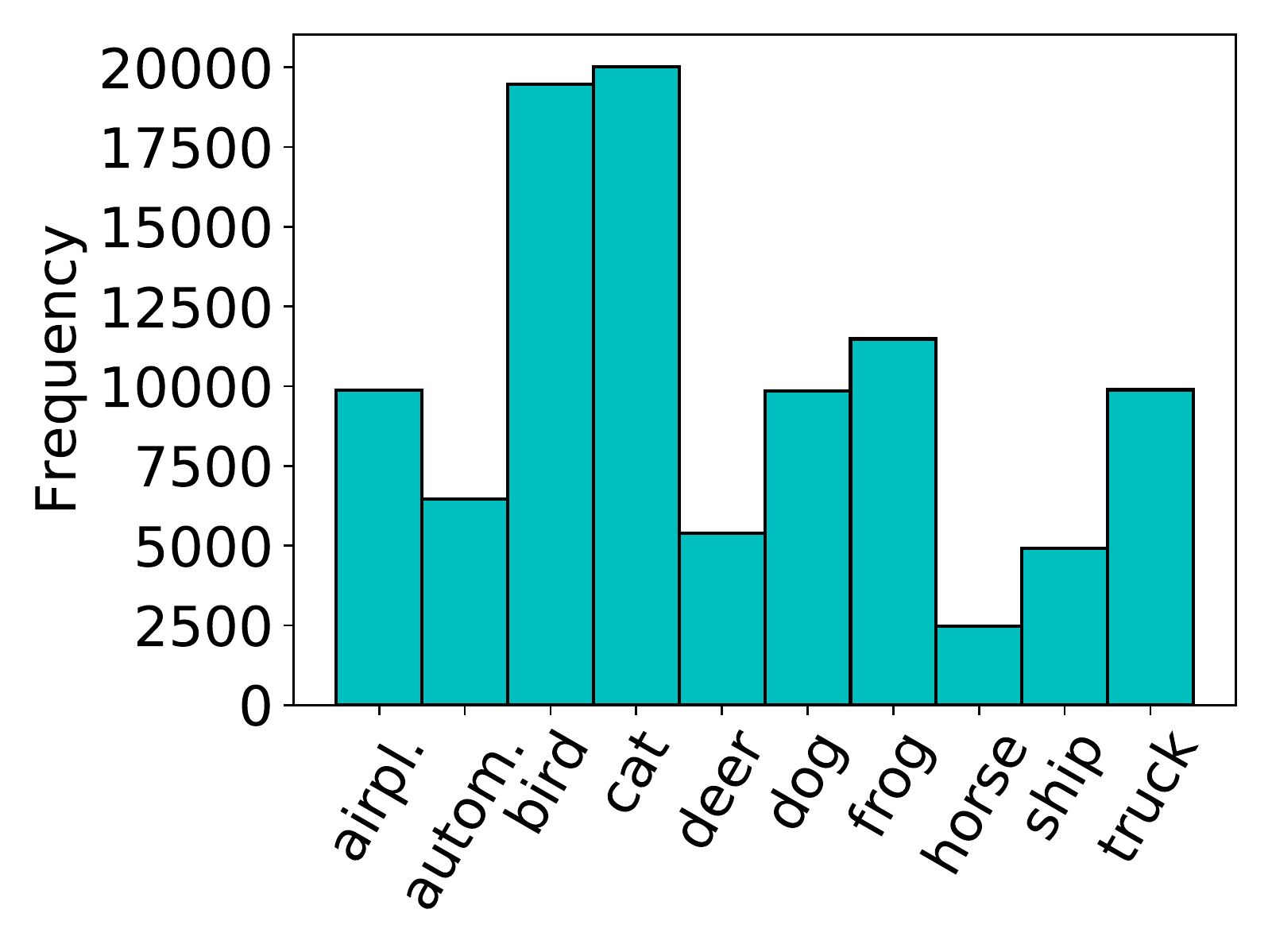}
	\qquad
		\begin{tabular}[b]{p{\dimexpr 0.18\linewidth-2\tabcolsep}
						    p{\dimexpr 0.12\linewidth-2\tabcolsep}
                          p{\dimexpr 0.18\linewidth-2\tabcolsep}}\hline
			\hfil Class name  & \hfil $Acc(F_\mathcal{V})$  & \hfil $Acc(F_\mathcal{A})$\\ \hline
			\hfil Airplane    & \hfil 95\% & \hfil 88\% (0.92 $\times$)   \\
			\hfil Automobile  & \hfil 97\% & \hfil 95\% (0.97 $\times$)\\
			\hfil Bird        & \hfil 92\% & \hfil 87\% (0.94 $\times$)\\
			\hfil Cat         & \hfil 89\% & \hfil 86\% (0.96 $\times$)\\
			\hfil Deer        & \hfil 95\% & \hfil 84\% (\textbf{0.88} $\times$) \\
			\hfil Dog         & \hfil 88\% & \hfil 84\% (0.95 $\times$)\\
			\hfil Frog        & \hfil 97\% & \hfil 90\% (0.92 $\times$)\\
			\hfil Horse       & \hfil 96\% & \hfil 79\% (\textbf{0.82} $\times$) \\
			\hfil Ship        & \hfil 96\% & \hfil 92\% (0.95 $\times$)\\
			\hfil Truck       & \hfil 96\% & \hfil 92\% (0.95 $\times$)\\ \hline
		\end{tabular}
		\caption{Histogram of \adv's transfer set constructed by querying CIFAR10-RN34 victim model with 100,000 ImageNet samples and per-class test accuracy for victim and surrogate models. The largest differences in per-class accuracies are in bold.}
		\label{fig:class-accuracy}
\end{figure}

When constructing $\{D_{\mathcal{A}}, F_\mathcal{V}(D_{\mathcal{A}})\}$, \adv might collect images that are irrelevant to the learning task or not close to \victim's training data distribution.
Moreover, \adv might end up having an imbalanced set, where observations for each class are disproportionate.
In this case, per-class accuracy of $F_{\mathcal{A}}$ might be much lower than $F_{\mathcal{V}}$ for classes with a few observations.
Figure~\ref{fig:class-accuracy} shows this phenomenon when $F_{\mathcal{V}}$ is CIFAR10-RN34.
For example, $Acc(F_{\mathcal{A}})$ is much lower than $Acc(F_{\mathcal{V}})$ in ``deer" and ``horse" classes.
When the histogram of $\{D_{\mathcal{A}}, F_\mathcal{V}(D_{\mathcal{A}})\}$ is checked, the number of queries resulting in these prediction classes are low when compared with other classes.
We conjecture that a realistic \adv might try to balance the transfer set by adding more observations for underrepresented classes or remove some training samples with less confidence values.


We further investigate the effect of a poorly chosen $D_{\mathcal{A}}$ by performing Knockoff nets against all $F_\mathcal{V}$'s using Diabetic5 as $D_{\mathcal{A}}$ (aside from Diabetic5-RN34).
We measure $Acc(F_{\mathcal{A}})$ to be between 3.9-41.9\%.
The performance degradation in this experiment supports our argument that the $D_{\mathcal{A}}$ should be chosen carefully by \adv.

\subsection{Access to In-distribution Data}
\label{sec:st-improvements}

A realistic \adv might know the task of the prediction API and could collect natural samples related to this task.
By doing so, \adv can improve its surrogate model by constructing $\{D_{\mathcal{A}}, F_\mathcal{V}(D_{\mathcal{A}})\}$ that approximates $F_\mathcal{V}$ well without being detected.

In section~\ref{sec:defense}, we observed that the higher the similarity between \adv's and \victim's training data distribution, the less effective our method becomes.
In the worst case, \adv has access to a large amount of unlabeled data that does not significantly deviate from \victim's training data distribution. In such a scenario, TNR values in Table~\ref{tab:detection-results} would clearly drop to 50\%. 
We argue that this limitation is inherent to all detection methods that try to identify out-of-distribution samples.

Publicly available datasets designed for ML research
as well as vast databases accessible through search engines and from data vendors (e.g. Quandl, DataSift, Axciom) allow \adv to obtain substantial amount of unlabeled data from any domain.
Therefore, making assumptions about \adv's access to natural data (or lack of thereof) is not realistic.
This corresponds to the most capable, and yet plausible, adversary model - one in which \adv has approximate knowledge about \victim's training data distribution and access to a large, unlabeled dataset.
In such a scenario, \adv's queries are not going to differ from a benign client, rendering detection techniques ineffective.
Therefore we conclude that against a strong, but realistic adversary, the current business model of prediction
APIs, which allow a large number of inexpensive queries cannot thwart model extraction attacks. 

Even if model extraction attacks can be detected through stateful analysis, highly distributed Sybil attacks are unlikely to be detected.
In theory, vendors could charge their customers upfront for a significant number of queries (over 100k), making Sybil attacks cost-ineffective.
However, this reduces utility for benign users and restricts the access only to those who can afford to pay. Using authentication for customers or rate-limiting techniques also reduce utility for benign users. Using these methods only slow down the attack and ultimately fail to prevent model extraction attacks.    

\section{Related Work}
\label{sec:related-work}

There have been several methods proposed to detect or deter model extraction attacks.
In certain cases, altering predictions returned to \client{s} has been shown to significantly deteriorate model extraction attacks: predictions can be restricted to classes~\cite{tramer2016stealing,orekondy2019defense} or adversarially modified to degrade the performance of the surrogate model~\cite{lee2018defending,orekondy2019defense}.
However, it has been shown that such defenses does not work against all attacks. Model extraction attacks against simple DNNs~\cite{juuti2019prada,orekondy2019defense,orekondy2018knockoff} are still effective when using only the predicted class.
While these defenses may increase the training time of \adv, they ultimately do not prevent Knockoff nets.

Other works have argued that model extraction defenses alone are not sufficient and additional countermeasures are necessary.
In DAWN~\cite{szyller2019dawn}, the authors propose that the victim can poison \adv's training process by occasionally returning false predictions, and thus embed a watermark in its model.
If \adv later makes the surrogate model publicly available for queries, victim can claim ownership using the embedded watermark.
DAWN is effective at watermarking surrogate models obtained using Knockoff nets, but requires that \adv's model is publicly available for queries and does not protect from the model extraction itself.
However, returning false predictions with the purpose of embedding watermarks may be unacceptable in certain deployments, e.g. malware detection.
Therefore, accurate detection of model extraction may be seen as a necessary condition for watermarking.

Prior work found that distances between queries made during model extraction attacks follow a different distribution than the legitimate ones~\cite{juuti2019prada}.
Thus, attacks could be detected using density estimation methods, where \adv's inputs produce a highly skewed distribution.
This technique protects DNN models against specific attacks using synthetic queries and does not generalize to other attacks, e.g. Knockoff nets.
Other methods are designed to detect queries that explore abnormally large region of the input space~\cite{Kesarwani2017model} or attempt to identify queries that get increasingly close to the classes' decision boundaries~\cite{quiring2018forgottensib}.
However, these techniques are limited in application to decision trees and they are ineffective against complex DNNs that are targeted 
by Knockoff nets.

In this work, we aim to detect queries that significantly deviate from the distribution of victim's dataset without affecting prediction API's performance.
As such, our approach is closest to the PRADA~\cite{juuti2019prada} defense.
However, we aim to detect Knockoff nets, which PRADA is not designed for.
Our defense exploits the fact that Knockoff nets uses natural images sampled from public databases constructed for a general task. 
Our defense presents an inexpensive, yet effective defense against Knockoff nets, and may have wide practical applicability. However, we believe that ML-based detection schemes open up the possibility of evasion, which we aim to investigate in future work.

\section{Conclusion}
\label{sec:conclusion}

We evaluated the effectiveness of Knockoff nets, a state-of-the-art model extraction attack, in several real-life scenarios. We showed that under its original adversary model described in~\cite{orekondy2018knockoff}, it is possible to detect an adversary \adv mounting Knockoff nets attacks by distinguishing between in- and out-of-distribution queries. While we confirm the results reported in~\cite{orekondy2018knockoff}, we also showed that more realistic assumptions about the capabilities of \adv can have both positive and negative implications for attack effectiveness. On the one hand, the performance of Knockoff nets is reduced against more realistic prediction APIs that do not return complete probability vector. On the other hand, if \adv knows the task of the victim model and has access to sufficient unlabeled data drawn from the same distribution as the \victim's training data, it can not only be very effective, but virtually undetectable. We therefore conclude that strong, but realistic adversary can extract complex real-life DNN models effectively, without being detected. Given this conclusion, we believe that deterrence techniques like watermarking~\cite{szyller2019dawn} and fingerprinting~\cite{lukas2019deep} deserve further study -- while they cannot prevent model extraction, they can reduce the incentive for model extraction by rendering large-scale exploitation of extracted models detectable. 

\subsubsection{Acknowledgements.} This work was supported in part by the  Intel (ICRI-CARS). We would like to thank Aalto Science-IT project for computational resources.

%
%
%
\bibliographystyle{splncs04}
\bibliography{main}
%




\end{document}